\documentclass[fleqn,10pt]{wlscirep}
\usepackage[toc,page]{appendix}
\usepackage{amsmath}
\usepackage{amssymb}
\usepackage{color}
\usepackage{rotating}
\usepackage{bigstrut}
\usepackage{multirow}
\usepackage{multicol}
\usepackage{tabularx}
\usepackage{colortbl}
\usepackage{hhline}
\usepackage{amsmath}
\usepackage{tabularx}
\usepackage{graphbox}
\usepackage{graphicx} 
\usepackage{booktabs} 
\usepackage{xparse}   

\usepackage{gensymb}
\usepackage{enumitem}
\usepackage[utf8]{inputenc}
\usepackage{graphicx}
\usepackage{grffile}
\usepackage[square,sort,comma,numbers]{natbib} 
\usepackage[export]{adjustbox}
\usepackage{footnote}

\usepackage{url}

\usepackage{url}
\usepackage[font={small,it}]{caption}
\usepackage{subcaption}
\usepackage[capitalise,noabbrev]{cleveref}
\usepackage{tabularx}
\newcolumntype{Y}{>{\raggedright\arraybackslash}X}

\graphicspath{ {} }

\newcommand{\TODO}[1]{}
\newcommand{\mario}[1]{}
\newcommand{\majid}[1]{}
\newcommand{\dominik}[1]{}
\newcommand{\change}[1]{\textcolor[rgb]{0,0,0}{#1}}
\newcommand{\scndchange}[1]{\textcolor[rgb]{0,0,0}{#1}}
\newcommand{\mjf}[1]{}

\title{Advanced Steel \change{Microstructural} Classification by Deep Learning Methods}
\author[1,2,4,*]{Seyed Majid Azimi}
\author[2,3]{Dominik Britz}
\author[2,3]{Michael Engstler}
\author[1]{Mario Fritz}
\author[2,3]{Frank Mücklich}

\affil[1]{Max Planck Institute for Informatics, Computer Vision and Multimodal Computing, Saarbr\"ucken, Germany}
\affil[2]{Material Engineering Center Saarland, Saarbr\"ucken, Germany}
\affil[3]{Saarland University, Chair of Functional Materials, Saarbr\"ucken, Germany}
\affil[4]{German Aerospace Center (DLR), Remote Sensing Technology Institute, We{\ss}ling, Germany (current affiliation)}
\affil[*]{seyedmajid.azimi@dlr.de}


\begin{abstract}
The inner structure of a material is called microstructure. It stores the genesis of a material and determines all its physical and chemical properties. While \change{microstructural} characterization is widely spread and well known, the \change{microstructural} classification is mostly done manually by human experts, which \scndchange{gives rise to uncertainties due to subjectivity}. Since the microstructure could be a combination of different phases \change{or constituents} with complex substructures its automatic classification is very challenging and only a few prior studies exist. \change{Prior works focused on} designed and engineered features by experts and \change{classified microstructures} separately from \change{the} feature extraction step. Recently, Deep Learning methods have shown \change{strong} performance in vision applications by learning the features from data together with the classification step. In this work, we propose a Deep Learning method for \change{microstructural} classification in the examples of certain \change{microstructural} constituents of low carbon steel. This novel method employs pixel-wise segmentation via Fully Convolutional Neural Networks (FCNN) accompanied by \change{a} max-voting scheme. Our system achieves 93.94\% classification accuracy, drastically outperforming the state-of-the-art method of 48.89\% accuracy.  Beyond the \change{strong performance of our method}, this line of research offers a more robust and first of all objective way for the difficult task of steel quality appreciation.
\end{abstract}
\begin{document}

\flushbottom
\maketitle
\thispagestyle{empty}

\section{Introduction}
\begin{figure}[!htb]
	\centering
	\includegraphics[scale=0.4]{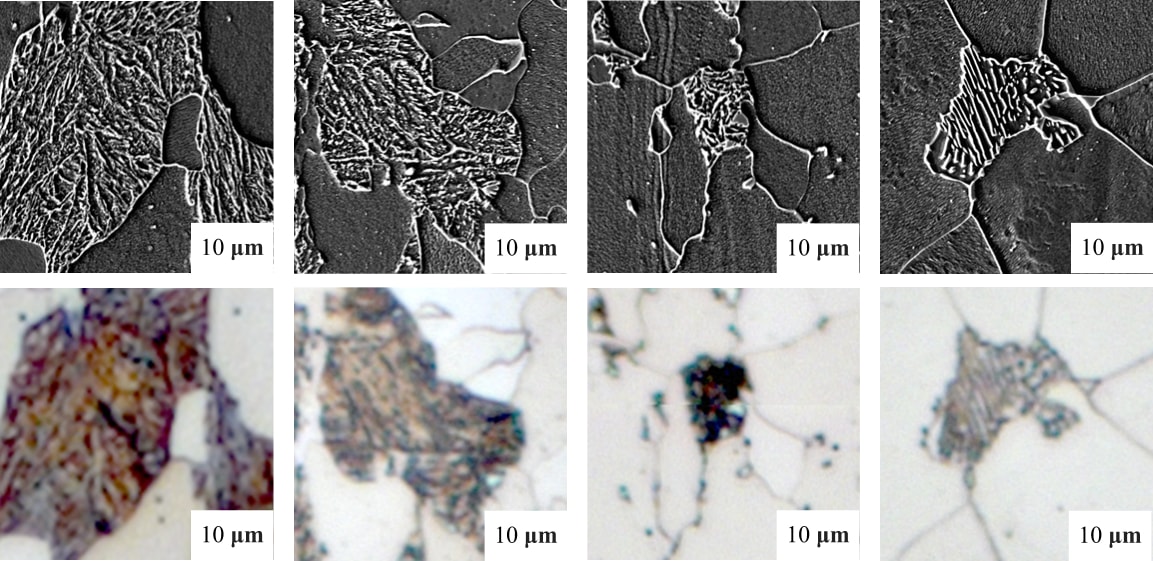}
	\caption{Some examples of different microstructure classes. In columns from left to right: martensite, tempered martensite, bainite and pearlite phases \change{or rather constituents} as \change{``objects''} (second-phase) have been illustrated. Ferrite is the matrix phase in \change{these} images, having the role of the background. The upper row contains images taken by Scanning Electron Microscopy (SEM) and lower row taken by Light Optical Microscopy (LOM).}
	\label{fig:objects}
\end{figure}

Steel is still one of the most important and \change{extensively used} \scndchange{classes} of materials because of its excellent mechanical properties while keeping costs low which gives a huge variety of applications\cite{tasan,khedkar}. The mechanical properties of steel are mainly determined by its microstructure\cite{harry} shown in Figure \ref{fig:objects}, so \scndchange{that} the performance of the material highly depends on the distribution, shape and size of phases in the microstructure\cite{ohser}. Thus, correct classification of these microstructures is crucial\cite{aarnts}. The microstructure of steels has different appearances, influenced by \change{a vast number of} parameters such as alloying elements, rolling setup, cooling rates, heat treatment and further post-treatments\cite{george}. Depending on how the steel is produced due to these parameters, the microstructure consists of different \change{constituents} such as ferrite, cementite, austenite, pearlite, bainite and martensite\cite{elena} shown in Figure \ref{fig:objects}.
    
This motivation leads us to use Deep Learning methods which are recently grabbing the attention of scientists due to their strong ability to learn high-level features from raw input data. Recently, these methods have been applied very successfully \change{to} computer vision problems\cite{vgg,lenet}. \change{They are based on artificial neural networks} such as Convolutional Neural Networks (CNNs)\cite{lenet}. They can be trained for recognition and semantic pixel-wise segmentation tasks. Unlike traditional methods in which feature extraction and classification are \change{learnt} separately, in Deep Learning methods, these parts are \change{learnt jointly}. \scndchange{The trained models have shown successful mappings from raw unprocessed input to semantic meaningful output}. As an example, \scndchange{Masci et al.}\cite{masci} used CNNs to find defects in steel. In this work, we show that Deep Learning can be successfully applied to identify \change{\change{microstructural}} patterns. Our method uses a segmentation-based approach based on Fully Convolutional Neural Networks (FCNNs) which is an extension of CNNs accompanied by a max-voting scheme to classify microstructures. Our experimental results show that the proposed method considerably increases the classification accuracy compared \change{to state of the art}. It also shows the effectiveness of pixel-based approaches compared \change{to} object-based ones in \change{microstructural} classification.

\section{Related Works} 

\scndchange{Based} on the instrument used for imaging, we can categorize the related works \scndchange{into} \change{Light Optical Microscopy} (LOM) and Scanning Electron Microscopy (SEM) imaging. High-resolution SEM imaging is very expensive compared with LOM imaging in terms of time and operating costs. However, low-resolution LOM imaging makes distinguishing microstructures based on their substructures even more difficult. 
Nowadays, the task of \change{microstructural} classification is performed by observing a sample image by an expert and assigning one of the microstructure classes to it. As experts are different in their level of expertise, one can assume that sometimes there are different opinions from different experts. However, thanks to highly professional human experts, this task has been accomplished so far with low error which is appreciated.
Regarding automatic \change{microstructural} classification, microstructures are typically defined by the means of standard procedures in metallography. \change{Vander Voort \cite{vander1} used Light Optical Microscopy (LOM) microscopy, but without any sort of learning the \change{\change{microstructural}} features which is actually  still the state of the art in material science for classification of microstructres in most institutes as well as in industry. His method defined only procedures with which one expert can decide on the class of the microstructure. Moreover, additional chemical etching\cite{beraha2} made it possible to distinguish second phases using different contrasts, however etching is constrained to empirical methods and can not be used in distinguishing various phases in steel with more than two phases. Nowadays, different techniques and approaches made morphological or crystallographic properties accessible\cite{5shreshta,7bhadeshia,8gerdemann,10international,ohser,britz,britz2}. Any approach for identification of phases in multiphase steel relies on these methods and aims at the development of advanced metallographic methods for morphological analysis purposes using the common characterization techniques and were accompanied with pixel- and context-based image analysis steps.}

\change{Previously, \scndchange{Velichko et al.}\cite{velichko2} proposed a method using data mining methods by extracting morphological features and a feature classification step on cast iron using Support Vector Machines (SVMs) - a well established method in the field of machine learning \cite{svm}. 
More recently, \scndchange{Pauly et al.}\cite{pauly} followed this same approach by applying on a contrasted and etched dataset of steel, acquired by SEM and LOM imaging which was also used in this work. However, it could only reach 48.89\% accuracy in microstructural classification on the given dataset for four different classes due to high complexity of substructures and not discriminative enough features.}

Deep Learning methods have been applied in object classification and image semantic segmentation for different applications. AlexNet, a CNN proposed by \scndchange{Alex Krizhevsky et al.,}\cite{alexnet} with 7 layers was the winner of ImageNet Large Scale Visual Recognition Challenge (ILSVRC)\cite{imagenet}in 2012 which is one of the most well-known object classification task challenges in computer vision community. It is the main reason that Deep Learning methods drew a lot of attention. AlexNet improved \change{the accuracy} ILSVRC2012 by 10 \scndchange{percent points} which was a huge increase in this challenge. VGGNet, a CNN \change{architecture} proposed by \scndchange{Simonyan et al.}\cite{vgg} has even more layers than AlexNet achieving better accuracy performance. \change{Fully Convolutional Neural Networks (FCNNs)} architectures, proposed by \scndchange{Long et al.,}\cite{fcnlong} is one of the first and well-known works to adapt object classification CNNs to semantic segmentation \change{tasks}. FCNNs and \change{their} extensions to the approach are currently the state-of-the-art in semantic segmentation on a range of benchmarks including Pascal VOC image segmentation challenge \cite{pascal} or Cityscape \cite{Cordts2016Cityscapes}.

Our method transfers the success of Deep Learning for segmentation tasks to the challenging problem of \change{microstructural} classification in the context of steel quality appraisal. It is the first demonstration of a Deep Learning technique in this context that in particular shows substantial gains over the previous state of the art.

\section{Review of recent Deep Learning techniques in computer vision}
Previous work in the context of steel \change{microstructural} classification relies on hand-designed features. Engineering high-level  features \change{ e.g.,} morphological \cite{pauly} features require complex feature extraction algorithms and a tedious \change{trial-and-error} process of finding good features in the first place. Deep Learning methods are capable of learning complex features from raw input data that turn out to also be superior across a wide range of application domains. These methods are inspired by neural networks and an ``end-to-end'' learning paradigm. Unlike classical methods, feature extraction and classification is done simultaneously in Deep Learning methods and optimized jointly. This is facilitated by composing complex, parameterized functions from simple, efficient, piece-wise differentiable building blocks. Therefore, training the whole system is facilitated by gradient descent on the parameters using a recursive gradient computation via the chain rule of differentiation (this is called back propagation algorithm \cite{backprop}). Among these Deep Learning methods, Convolutional Neural Networks (CNNs) and their modification, Fully Convolutional Neural Networks (FCNNs) have been shown particularly successful for image classification and segmentation.

CNNs were developed originally for vision tasks. The basic concept of CNNs dates back to \change{1980}\cite{Fukushima,lenet}, but due to the increase in computational resources, available data and better learning algorithms, they have recently attained \change{a new level of quality} across a range of domains. \change{Therefore, CNNs are the state-of-the-art} approaches in many vision applications. As images are of high dimensionality applying traditional neural networks with fully connected neurons to process visual data would lead to a huge number of trainable parameters. Furthermore, they are not able to exploit the natural structures in the images like correlation among neighboring pixels and stationary image statistics.
Typical Convolutional Neural Networks consist of multiple, repeating components that are stacked in layers: convolution, pooling, fully connected and classifier layers.

\textbf{Convolution Layer} convolves the input data with a linear convolution filter as shown in  Equation \ref{eq:convolution}:
\begin{equation}\label{eq:convolution} 
(h_{k})_{ij} = (W_{k}\ast x)_{ij} + b_{k}
\end{equation}
\change{w}here $k=1,\dots,K$ is the index of \change{the} $k$-th feature map in convolution layer and $(i, j)$ is the index of neuron\change{s} in \change{the} $k$-th feature map and $x$ represents \change{the} input data. $W_{k}$ and $b_{k}$ are trainable parameters (weights) of linear filters (kernel) and bias for neurons in \change{the} $k$-th feature map respectively. $(h_{k})_{ij}$ is \change{the }value of the output for the neuron in \change{the} $k$-th feature map with position of $(i, j)$. The spatial 2D convolution operation between \change{the} input data and \change{the} feature map has been represented by ``$\ast$''.

\textbf{Pooling Layer} is a nonlinear down-sampling layer which either takes \change{maximum} or average values in each sub-region of \change{the} input data. \change{Today's CNNs typically employ a maximum pooling called ``max-pooling'' layer in order to achieve invariance to small shifts in the feature maps.}

\textbf{Fully-connected Layer} is a classic neural network layer where the features of the next layer are a linear combination of the features of the previous layer as shown in Equation \ref{eq:fullyconnected}:
\begin{equation}\label{eq:fullyconnected} 
y_{k}= \sum_{l}W_{kl}x_{l} + b_{k}
\end{equation}
\change{w}here $y_{k}$ represents the $k$-th output neuron \change{and} $W_{kl}$ is the $kl$-th weight between $x_{l}$ and $y_{k}$.

\textbf{Activation Function} usually follows a pooling or fully connected layer and introduces a nonlinear activation operation like \change{a} sigmoid or rectified linear unit (ReLU). \change{The} ReLU function $relu(x)=\max(0,x)$ are most common as the gradient is piece-wise constant and does not vanish for high activations (in contrast to \scndchange{ e.g.,} sigmoid).  \\
\textbf{Classifier Layer} is the last layer of the network and computes a class posterior probability of \change{the} input image. The most widely used classifier layer in CNNs is \change{the} softmax function. The vector of real values between $(0, 1)$ generated by this function denotes a categorical probability distribution shown by Equation \ref{eq:softmax} for \change{the} $j$-th class and an input vector $X$.
\begin{equation}\label{eq:softmax} 
P(y=j|X;W,b)= \frac{\exp^{X^{T}W_{j}}}{\sum_{k=1}^{K}\exp^{X^{T}W_{j}}}
\end{equation}
\textbf{Loss Layer} is used to measure the difference between true class labels and corresponding predicted class labels. \change{The most common} loss layer for classification is the cross-entropy loss. The \change{c}ross-entropy loss \change{is shown} in Equation \ref{eq:crossentropy}:
\begin{equation}\label{eq:crossentropy} 
\text{Cross-entropy Loss Function}= -\sum_{x}{P^{'}(x) \log{P(x)}}  
\end{equation}
in which \change{the} softmax classifier ($P(x)$) is minimizing the cross-entropy between the ``true'' one-hot encoded distribution of data($P^{'}(x)$) and the predicted class probabilities.

\textbf{CNN Training} is done end-to-end which means without separation between \change{the} feature extraction and \change{the} classification step. They receive raw input data \change{ e.g.,} image pixels, to produce semantic outputs. CNNs are trained to learn the hidden pattern in the data \change{by using a training set}. \change{A} loss function $\mathcal{L}$ measures how far the outputs of the network from the correct output is. This optimization problem is solved by gradient descent\cite{lenet} and a well-known process called Back-Propagation\cite{lenet} to propagate the loss function gradient to previous layers. The gradient descent is performed based on \change{the} loss function gradient $\frac{\delta \mathcal{L}}{\delta w_{ij}}$. Then, the weights are adapted in order to decrease the loss function output by modifying into the direction opposite to the direction of increasing gradient with a step size (learning rate) represented by $\eta$ in Equation \ref{eq:gradientdescent}. \change{Learning rate does not have a unit. It is a user defined parameter. It determines how much gradient of loss function is applied to update  weights in each back-propagation step during CNN training phase.}
\begin{equation}\label{eq:gradientdescent}
w_{ij}^\textbf{new} = w_{ij}^\textbf{old}-\eta \frac{\delta \mathcal{L}}{\delta w_{ij}}
\end{equation}
\textbf{Dropout Layer\cite{dropout}} is a technique to improve \change{the} generalization power of CNNs by randomly ignoring (dropping) neurons and their corresponding parameters from the network architecture only during \change{the} training phase.

As annotated SEM images are rare, most likely \change{a} training network on such dataset will lead to overfitting to noise present in the training set. To address this problem, networks which have already been trained using large training datasets like ImageNet are trained on the new dataset. This trick is known as ``Transfer Learning'' or ``fine tuning''. Using this technique, we can initialize the weights more efficiently. Therefore, it can be assumed that the network is already close to the best local minimal solution and needs far less training data to converge. From pre-trained CNNs, features can also be extracted without \change{fine tuning}. In this case, the network is not fine tuned. Instead, output of fully connected layers before \change{the} classification layer is considered as feature vector. These features known as DeCAF\cite{decaf}  and can be classified with \scndchange{ e.g.,} SVMs.
Another trick in case of utilizing \change{small} datasets is to artificially enlarge the training set \scndchange{ e.g.,} by flipping or rotating while preserving the class labels. This trick is known as ``Data Augmentation''.

\textbf{Network Architectures:}
In the following, we describe the three specific CNN architectures (with increasing depth) that we use for classification:
     {\it CIFARNet} is a CNN with \change{three} convolutional and max-pooling layers with two fully-connected layers. It is a modified version of LetNet\cite{lenet} proposed by \scndchange{Lecun et al.,} containing 431K parameters to which \change{the} ReLU activation layer and dropout layer (will be described in the following) have been added.

     {\it AlexNet}\scndchange{,} proposed by \scndchange{Alex Krizhevsky et al.,}\cite{alexnet} is a deep CNN with 60 million parameters.

AlexNet is deeper than CIFARNet. It has \change{eight} layers including \change{five} convolutional, \change{three} max-pooling and \change{three} fully-connected layers.
{\it VGGnet} \change{was} proposed by \scndchange{Simonyan et al.}\cite{vgg} \change{and}
it is even deeper with 13 convolutional, \change{five} pooling and \change{three} fully-connected layers known as VGG16 with 138 million parameters. Another version called VGG19 \change{has} 19 layers.
VGG19 network was able to achieve empirically better performance than CIFARNet and AlexNet.
In VGGnet, \change{a} 3x3 convolution kernel size was applied which resulted in \change{fewer} parameters, but with the same support.

\subsection{Fully Convolutional Neural Networks (FCNNs)}
While CNNs have been shown successful for image classification, we are also interested in predicting a set of semantic classes for each pixel which is called semantic segmentation. CNNs can be extended to perform this task by removing the fully-connected layers and ``sliding'' the CNN over a larger image.  However, the resulting semantic segmentation would not quite be of the same resolution as the original image. Therefore, \scndchange{Long et al.}\cite{fcnlong} proposed Fully Convolutional Neural Networks(FCNNs) with an additional up-sampling layer.

\textbf{\change{Up-sampling} layer} can be achieved by simple bilinear interpolation (which is differentiable). Today's \change{FCNNs} learn also the \change{up-sampling} kernel which is parameterized as shown in Equation \ref{eq:upsamp}:

\begin{equation}\label{eq:upsamp} 
y_{ij} = \sum_{\alpha,\beta=0}^{1}|1-\alpha -\{i/f\}|\;\;|1-\beta-\{i/j\}|\;\;x_{\lfloor i/f\rfloor+\alpha,\lfloor j/f\rfloor+\beta,}
\end{equation}

where $y$ and $x$ are \change{the} input and output of \change{the} up-sampling layer, and $f$ and \change{${\alpha, \beta}$} stand for the up-sampling factor and pixel fraction parts, respectively. One can consider \change{an} up-sampling layer with factor $f$ as a convolution operation, but instead of \change{an} integer stride it has \change{a} fractional input stride of $1/f$. 

To capture different information about the input data by preserving the dimensions of input feature maps, we can fuse different pooling layers together known as ``skip layers''. FCNNs can be seen as an encoder-decoder system in which convolutional layers do encoding of input images by extracting high-level learned feature\change{s} and deconvolution layers do decoding on these features to present semantic segmentation of the input.
As image semantic segmentation datasets are typically small, these networks are frequently pre-trained on object recognition datasets and then fine-tuned to perform the segmentation task.

\section{Methods}
In this \scndchange{section}, we describe our methods to classify microstructures in steel. First, we applied Deep Learning methods to classify each cropped steel object \change{as illustrated in Figure \ref{fig:objectcnn}} from SEM or LOM images which we call object-based microstructural classification. Then we explain our main methods which classify each pixel as one of \change{a} microstructure class and then we classify each object by considering the classes of pixels inside the object. \change{To avoid misunderstandings, ``substructure'' and ``texture'' is used equally, owed the different current languages in material science and computer vision.} \change{All experimental protocols have been approved by Material Engineering Center Saarland (MECS) and Max Planck Institute for Informatics (MPI) institutes.} The methods were carried out in accordance with the relevant guidelines and regulations.    

\subsection{Object-based Classification of Microstructures with CNNs}

The first approach we \change{used, was the classification of}  microstructures (second/dual phases) by CNNs. However, CNNs work on images of fixed size. Therefore, we normalized the images for each structure by cropping  objects from the images,  resized \change{them} to a fixed size and then applied the CNN classifier.

\begin{figure}[!htb]
	\centering
	\includegraphics[width=\textwidth]{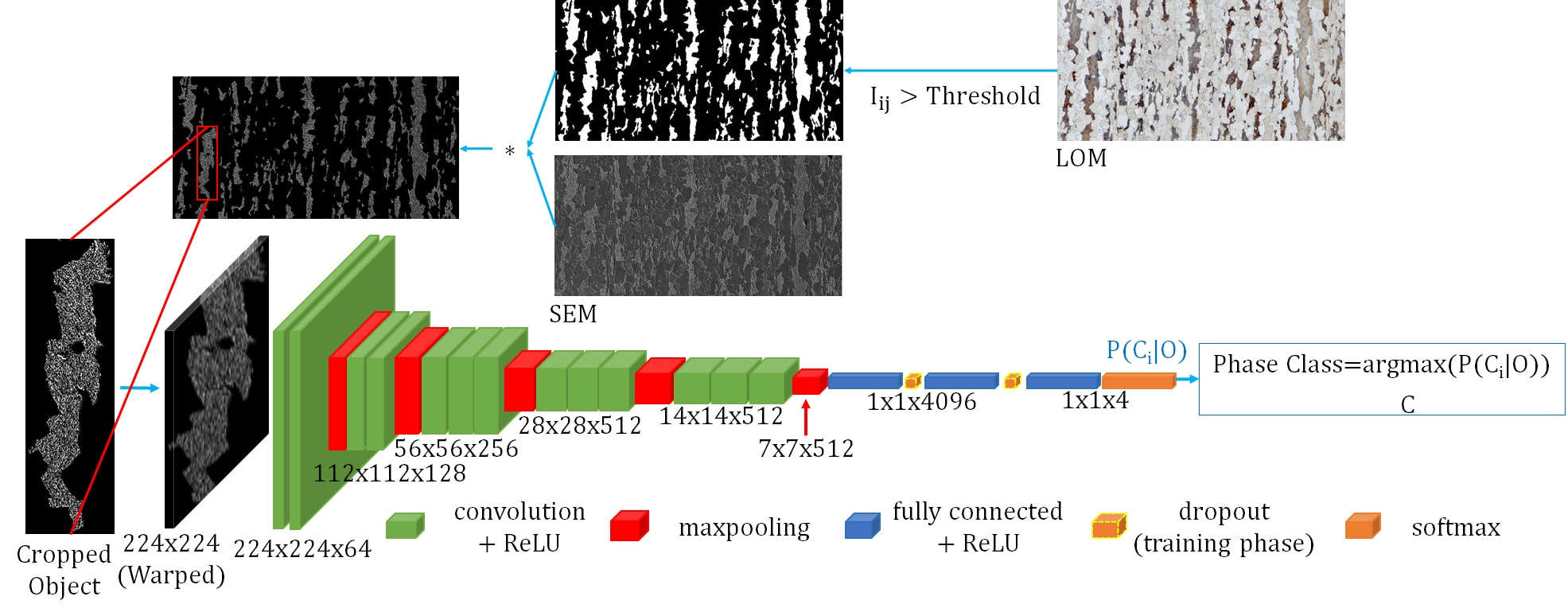}
	\caption{Workflow of \change{o}bject-based classification approach using CNNs. In this figure one object in \change{the} SEM image is cropped and classified using \change{a} trained CNN. 224x224\change{ px} is the fixed input size of \change{the} VGG16 network and ``C'' stands for class.}
	\label{fig:objectcnn}
\end{figure}

\textbf{\change{Masking:}} \change{By applying a user-defined threshold on pixel intensity in LOM image, binary segmented LOM as shown in Figure \ref{fig:objectcnn} can be computed. Corresponding SEM image which has already been registered with LOM image from the same area are masked using the mentioned binary mask. In the resulting masked SEM all of matrices (here ferrite) are masked as illustrated in Figure \ref{fig:objectcnn}. In addition, by using this mask, each constituent can be localized.}

\textbf{Cropping:} \change{Using location information obtained from the binary segmented LOM, each constituent (object) in the masked SEM image is cropped and separated from the main image as shown in the part of ``cropped object'' in Figure \ref{fig:objectcnn}}.\change{ Hence, the cropping operation is systematic.
These operations have been described with more details in \scndchange{Britz et al.}\cite{britz3}} 

\textbf{\change{Warping:}} \change{Due to fully-connected layers in CNNs, input images should be warped to a fixed size \scndchange{e.g., }224x224px for VGG16 network. The cropped objects should be then warped to a fix size.}

\textbf{Classification:} \change{Each warped constituent (object) image is then given to CNN as input to be classified. In this way, the system performance is comparable to the previous work of \scndchange{Pauly et al.}\cite{pauly} As the objects have been automatically warped and split randomly into training and test set, their distribution can be assumed identical independently distributed.} We considered three possible techniques of using CNNs for image classification: (I) full-training (from scratch) CNNs, (II) fine-tuning CNNs and (III) DeCAF features with SVM classifier. Since in this case, classification was done directly based on the object input image, we refer to this system as \change{the} object-based CNN \change{microstructural} classification.

The (I) technique is a network which was trained with random initialized parameters. In this method we are free to choose \change{the} size of \change{the} input image. 
The (II) strategy was using \textit{transfer learning}. However, in this case, we were limited to the input size of \change{the} CNN.
    
     The first and second strategy is illustrated in Figure \ref{fig:objectcnn}. The third strategy is similar except that classification is done by SVMs rather than \change{a} softmax layer. The output of the object-based CNN before classification is $P(C_{i}|O)$, where $C_{i}$ is the class of each phase, $O$ stands for the observation or the input image, and $P$ is the posterior probability that the input image belongs to class $i$. Classification is performed by choosing the class with the highest probability.
     
\subsection{Segmentation-based Classification of Microstructures with FCNNs}
\begin{figure}[!htb]
		\centering
		\includegraphics[width=\textwidth]{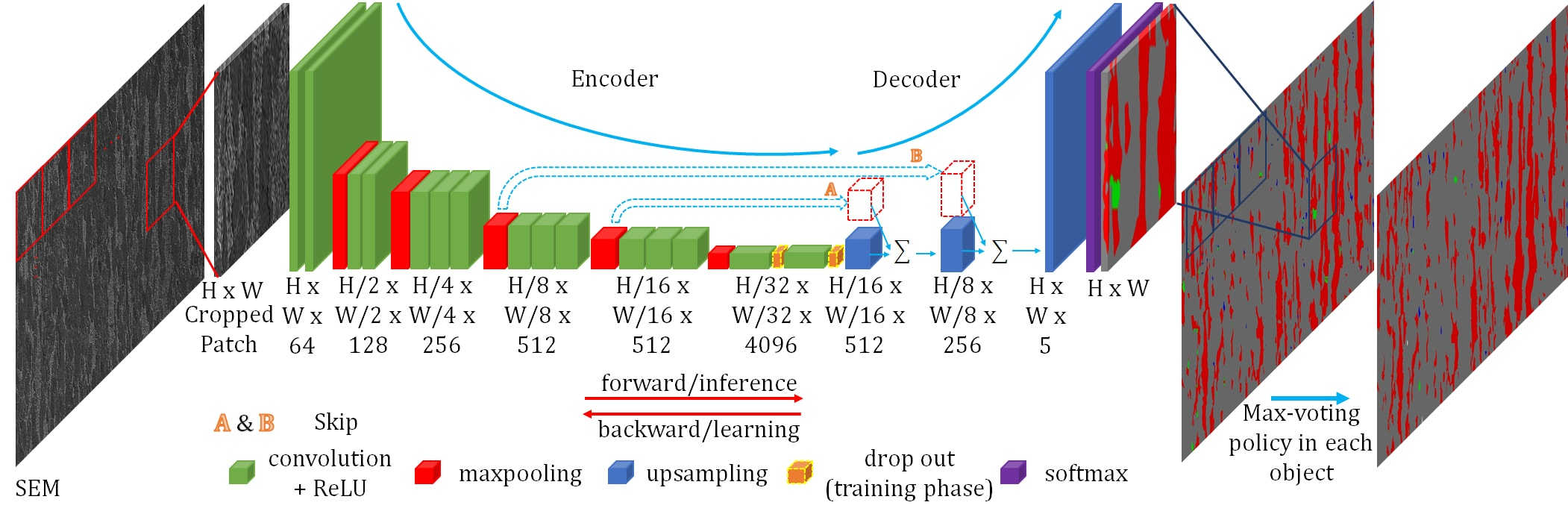}
		\caption{Workflow of max-voted segmentation-based \change{microstructural} classification approach using FCNNs (MVFCNN). \change{In this figure, the input image is an SEM image. The input image is first cropped. Then cropped patches are given to \change{the} FCNN part of \change{the} MVFCNN algorithm. The segmented crops are stitched together by FCNN. In the last step, \change{the} Max-voting policy \change{is} applied on the resulting stitched image. The max-voted output is used to classify microstructure objects.} H and W represent height and width and \change{the} third number is \change{the} number of \change{the} feature map.}
		\label{fig:maxvotfcnn}
	\end{figure}

 Resizing of each cropped object image to a fixed size, as required in the object-based CNN approach, could destroy valuable information related to the phase texture by heavy distortion. On the other hand, pixel-wise classification (segmentation) can work with any image size. Thus, we propose \change{a} SEM or LOM image segmentation-based \change{microstructural} classification approach using \change{a} FCNN and max-voting scheme to classify each object. The processing pipeline for this approach is illustrated in Figure \ref{fig:maxvotfcnn}. We refer to this approach as max-voted FCNN (MVFCNN). Using this method, SEM or LOM images are classified pixel-wise. In our experiments, we used \change{a} network architecture \change{proposed by Long et.al.}\cite{fcnlong}. The architecture is almost the same as VGG16, except \change{a converting from  fully-connected layers to convolutional ones and up-sampling layers plus skip layers}. They used skip layers to fuse coarse and local appearance information to improve the semantics. By using no skip layer, the network is called FCN-32s, with skip layer A  and with A and B together denoted in Figure \ref{fig:maxvotfcnn} network is called FCN-16s and  FCN8s respectively.
We used cropped raw SEM or LOM images as input to FCNNs. \change{For pixel-wise \change{microstructural} classification, we can not use the original image sample due to the big size of samples(~7000x8000 px) and GPU memory limit. Therefore, original image should be cropped into small patches and each patch should be segmented separately. Patch cropping is done using sliding window technique \scndchange{(e.g., Dalal et al.\cite{dalal})}. We consider a window of HxW size illustrated in Figure \ref{fig:maxvotfcnn}. This window slides over the original image with a horizontal and vertical step sizes. We call these step sizes as stride parameters. In each step, the original image is cropped using the sliding window. Then the cropped patch is given to FCNN as input image. We do this operation for all of patches to cover the whole input image. In other words, the whole original input image will be given as separate input patches to the network.} The \change{maximum} size of the cropped images is determined by \change{the} GPU memory.

The output of FCNN is a 3D matrix with the number of channels equal to the number of classes. Each pixel in this matrix has a value representing the score confidence or posterior probability (output of the softmax classifier function) for the corresponding microstructure class $C_{i}$. The pixel-wise classification step is then performed by choosing the class for each pixel with the highest posterior probability. Afterwards, all of \change{ the} segmented patches belonging to \change{the original} input image are stitched together as illustrated in Figure \ref{fig:maxvotfcnn}. In order to classify objects (microstructures) rather than pixels, \change{a} max-voting policy is applied on each object, assigning it to the class of the majority of the pixels. In other words, \change{the microstructure class with the maximum classified pixels inside an object will be assigned to that object.} The location information of objects is obtained using \change{a} binary LOM \change{image}. The motivation for using this aggregation step was that \change{by} using stitched patches we can decide the class of each object instead of \change{only a} part of it.
 
\subsection{Implementation Details}
In order to train and test CNNs and FCNNs, Caffe \cite{caffe} framework and a K40m NVIDIA GPU was used. \change{Caffe framework is a library in which most of \change{the} fundamental layers of neural networks have been implemented efficiently by C++ and Cuda programming languages.}

\subsubsection{Training object-based CNN:} All of the cropped object images were resized to 224x224\change{px} which is the fixed input size of VGG16. We also considered this size of input for training networks from scratch. We used a fixed learning rate of 0.001, a momentum of 0.9 and weight decay of 0.004 in stochastic gradient descent algorithm. The training iterations continue until the training loss reached a plateau. For training CIFARNet from scratch, the standard deviation of Gaussian noise for initial random weights for the first convolutional layer is 0.0001 and for the rest is 0.01. For \change{fine tuning}, \change{a} pre-trained VGG16 network was used. We initialized the last fully-connected layer with random Gaussian noise with standard deviation of 0.01. The learning rate of 0.0001 (chosen on validation set) was used to train CIFARNet and VGG16 respectively. 

\subsubsection{Using pre-trained extracted features (DeCAFs):}To classify DeCAF features using SVMs, we trained a multi-class SVM with RBF kernel with extracted features from pre-trained VGG19\cite{vgg} network. \change{In VGG19 architecture, a fully-connected layer before the  classification layer (size of 1 x 1 x 4096px) was considered as the feature vector. Therefore, the feature vector is a 1x1x4096 dimension vector.}

\subsubsection{Training MVFCNN:} In training MVFCNN, we used learning rate of $10^{-10}$, $10^{-11}$ and $3*10^{-12} $ to train FCN-32s, FCN-16s, and FCN-8s, respectively. \change{A} bigger learning rate causes the training loss to explode. The momentum of $0.9$ with weight decay of $5*10^{-4}$ was considered. Regarding input images, patches were cropped with $1000\times1000$ px size with \change{a} batch size of 1, due to memory issues. We first trained \change{a} FCN-32s model, and then added \change{a} skip layer (FCN-16s) and fine-tuned it. Afterwards, another skip layer (FCN-8s) was added to fine-tune the final model. Direct training of FCN-8s \change{gave} worse results. \change{A} pre-trained FCN-32 model was trained with \change{an} ImageNet dataset. The network was trained with 7000 iterations for $\sim 4.5$ days. The inference time for a 1000x1000 px input image was $\sim 600$ ms.

\subsubsection{Class Balancing and Data Augmentation} In order to address the problem of class unbalance in the dataset, in \change{the} MVFCNN method cropping was carried out for different classes with different stride \change{(step size)} parameters in horizontal and vertical directions which in the end all of classes had the same number of patches\change{ i.e.} the class with least number of images had smaller stride than the class with the largest number of training images. \change{Stride parameters are the parameters determining the steps using which patches are cropped. For example if \change{the} horizontal stride is 100px, it means when a patch is cropped, the next patch will be cropped with \change{a} 100px step. The same goes for the vertical direction.} \change{In our experiments, we chose different stride parameters corresponding to the number of images for each class. Stride parameters in horizontal and vertical directions were chosen the same. For an example, an image of 7000x8000 px includes $7*8=56$ patches with 1000x1000 px size stitched together. By using a stride parameter of 100 px, one can produce $61*71=4331$ patches each with the size of 1000x1000 px.} Resulting cropped patches were also rotated \change{by} 90$^{\circ}$, 180$^{\circ}$ and 270$^{\circ}$ to augment \change{the} dataset. In this case, the number of the training images \change{can} increase three times.

\section{Results}
\subsection{Dataset} 
In our experiments, we use a steel image dataset \cite{pauly} provided by Material Engineering Center Saarland (MECS) in Saarbr\"ucken, Germany.

\begin{figure}[!htb]
	\centering
	\begin{subfigure}{1\textwidth}
		\includegraphics[scale=0.25]{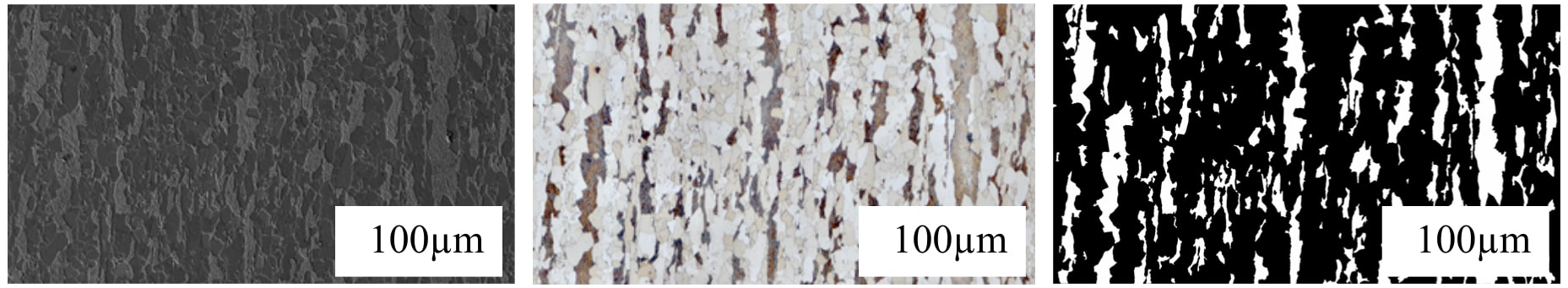}
		\caption{}
	\end{subfigure}
	\begin{subfigure}{1\textwidth}
		\includegraphics[scale=0.25]{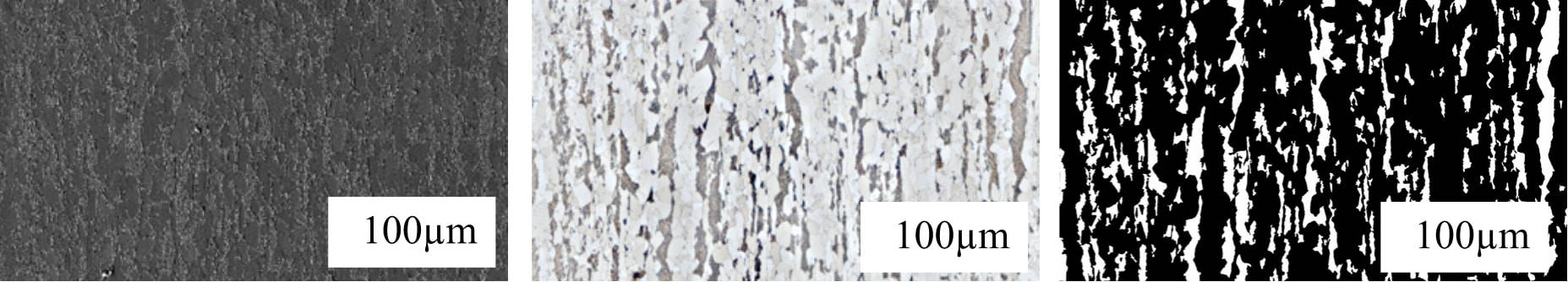}
		\caption{}
	\end{subfigure} 
	\begin{subfigure}{1\textwidth}
		\includegraphics[scale=0.25]{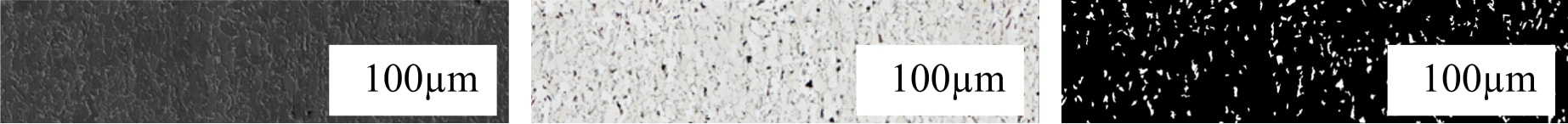}
		\caption{}
	\end{subfigure}
	\begin{subfigure}{1\textwidth}
		\includegraphics[scale=0.25]{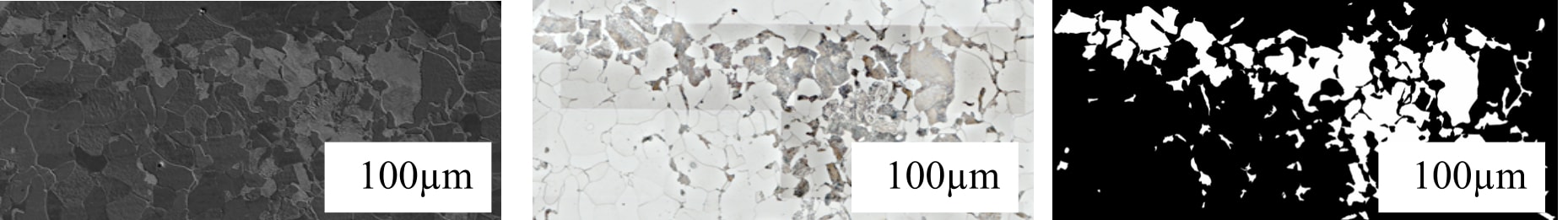}
		\caption{}
	\end{subfigure} 
	\caption{SEM and LOM example images for each microstructure class with ferrite as matrix and with diameter of up to 100 $\mu m$. The columns show SEM, LOM and segmented LOM images from left to right. Size ratios of samples have been preserved. The sub-images are corresponding to (a) martensitic, (b) tempered martensitic, (c) bainitic and (d) pearlitic microstructures.}
	\label{fig:dataset}
\end{figure}
The dataset contains registered images of steels taken by LOM and SEM  \change{and the} corresponding binary LOM images. The steel samples were first polished and etched according to \scndchange{Britz et al.}\cite{britz2} and Pauly et al.\cite{pauly} \change{In total,} 21 LOM and SEM images with an average size of 7000x8000px \change{were available}. LOM and SEM images for each microstructure class are shown in Figure \ref{fig:dataset}. \change{For ground truth, the procedure according to \scndchange{Britz et al.}\cite{britz3} was applied. Afterwards a group of material experts and metallographers assign the objects of the second phase to the mentioned phases/constituents according to Aarnts et al\cite{aarnts}. In other words, the assignment of the images to each microstructural class (ground truth) has been made by material science experts and metallographers. For example, if one SEM image contains martensitic and ferritic constituents, a group of material science experts have assigned martensite and ferrite labels to the ``objects'' or rather grains which the mentioned sample contains. Informed consent has been obtained from all material expert co-authors from MECS.}

\change{The microstructures contain a ferritic matrix as first phase and a pearlitic, martensitic or bainitic microstructures as second constituent. Therefore, objects are also referred to as two-phase  steel images with two constituents.} 21 images are distributed into martensitic (11 samples), tempered martensitic (2 samples), bainitic (4 samples) and pearlitic (4 samples). \change{A further subdivision for the bainite (upper, lower, granular etc. according to \scndchange{Zajac et al.}\cite{zajac}) will be considered in the next step after this proof of concept.}
The dataset split is image-based with 11 training images and 10 test images which results in 2831 training and 2262 test objects. \change{It should be mentioned that in the test set, one of two bainite samples is a bainitic sample which was normalized afterwards and therefore, it shows a different appearance despite being in the same class.}

\subsection{Task Definition and Metrics}
\begin{figure}[!htb]
	\centering
		\includegraphics[width=0.7\linewidth]{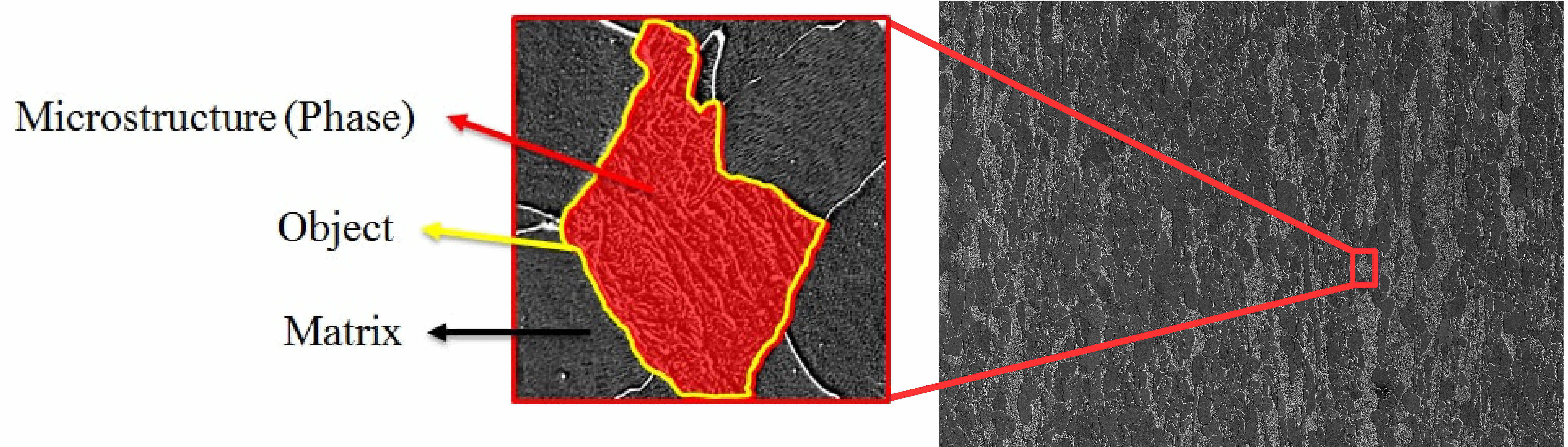}
	\caption{\change{SEM image of a steel microstructure}. The right image is an original SEM image and the left one shows a magnified object. In this example, the substructure (microstructure) is \change{martensitic within a ferritic matrix (background/first phase). The foreground (second/dual phase/constituent) will be considered as ``object''.}}
	\label{fig:problem}
\end{figure}

In the \change{microstructural} classification problem, the goal is to classify \change{constituents (object)} inside steel images based on microstructure classes. \change{These objects are grains within the microstructure which can be considered as ``background'' (matrix (ferrite)) and ``foreground'' (second constituent in the present microstructure) according  to Figure \ref{fig:problem}}. The substructure of such an object can be seen as texture \change{ - which should not be confused with the meaning of the term ``texture'' in Material Science -  which will be classified with the correct label in the present work.}
Therefore, the more objects \change{that} are classified correctly, the more accurate the system is.

In addition to the classification task, we also evaluated the semantic segmentation performance using the metrics in Equation \ref{eq:pixacc}, \ref{eq:meanacc}, \ref{eq:meaniu}, and \ref{eq:fmeaniu}. For more information, the reader is encouraged to read the paper by Long et .al.\cite{fcnlong} In the following, $n_{ij}$ is \change{the number of the} pixel of class $i-\text{th}$ predicted as class $j$, $n_{cl}$ is the number of different classes and $t_i=\sum_{j}n_{ij}$ is the whole pixel number of class $i$:
\change{
\begin{itemize}
	\item pixel accuracy: \begin{equation}\label{eq:pixacc} \frac{\sum_{i}n_{ii}}{\sum_{i}t_{i}} \end{equation}
    \item mean accuracy: \begin{equation}\label{eq:meanacc} \frac{1}{n_{cl}}\sum_{i}\frac{n_{ii}}{t_i} \end{equation}
	\item mean intersection over union (mean IU): \begin{equation}\label{eq:meaniu} \frac{1}{n_{cl}}\sum_{i}\frac{n_{ii}}{t_i+\sum_{j}n_{ji}-n_{ii}} \end{equation}
    \item frequency weighted intersection over union (frequency weighted IU): \begin{equation}\label{eq:fmeaniu} (\sum_{k}t_{k})^{-1} \frac{\sum_{i}t_{i}n_{ii}}{t_i+\sum_{j}.n_{ji}-n_{ii}} \end{equation}
\end{itemize}
}
In \change{microstructural} classification via \change{a} pixel-based MVFCNN approach, segmented objects are classified with the class that \change{the} majority of those pixels vote for (max-voting). \change{Then the evaluation is carried out by enumerating correct classified microstructure objects.} After this step, the results of object-based and pixel-based methods are comparable with each other as well as with the state of the art.

\subsection{Evaluation}
In Table \ref{tab:1}, comparable results of our experiments with object-based CNN and pixel-based MVFCNN \change{are} presented using SEM images. \change{Also} the performance of the \change{previous state-of-the-art method of \scndchange{Pauly et al.}\cite{pauly} with an accuracy of 48.89\% is shown}. \change{In their work, they enumerated the correctly classified microstructural objects by data mining tasks using the same dataset. Hence, the performance of our system both object-based CNN and pixel-based MVFCNN are comparable with theirs.} Instead, from-scratch trained CIFAR-Net is able to outperform this method by achieving 57.03\% accuracy. Using pre-trained VGG19-Net features with \change{a} SVM classifier and \change{a} fine tuned VGG16-Net \change{we} even \change{achieve a} better performance of 64.84\% and 66.50\%, \change{respectively}. All of these methods apply object-based algorithms. However, Table \ref{tab:1} shows \change{that microstructural} classification using pixel-based methods can achieve \change{a} considerably higher accuracy of 93.94\% accuracy which shows \change{that the} pixel-based classification of microstructures is \change{a very promising approach}.

\begin{table}[!htb]
	\centering
	\caption{\change{Microstructural} classification results using object-based CNN and pixel-based MVFCNN approaches. The results show \change{that} object-based classification approaches improve over prior work at most by around 18 \scndchange{percent points}. \change{The} pixel-based approach has even better performance by around 45 \scndchange{percent points} improvement.}
	\label{tab:1}
	\begin{tabular}{lccc}
    \toprule
		Method & Type &\begin{tabular}[c]{@{}c@{}}Training\\ Strategy\end{tabular} & Accuracy \\ \midrule
        \scndchange{Pauly et al.}\cite{pauly}  & object-based  & -- &48.89\%  \\ \midrule
		CIFAR-Net & object-based &  from scratch &57.03\% \\
        pre-trained VGG19-Net Features + SVM & object-based & -- & 64.84\% \\
		\begin{tabular}[c]{@{}c@{}}VGG16-Net\end{tabular} & object-based & \change{fine tuning} & 66.50\%  \\ \midrule 
     MVFCNN & pixel-based & \change{fine tuning} & \textbf{93.94\%}\\ \bottomrule
	\end{tabular}
\end{table}

In Table \ref{tab:2}, the effect of data augmentation and fine tuning using SEM images in \change{the} MVFCNN approach have been depicted. The results shows \change{that fine tuning} and using balanced augmented training data achieves the best results of 93.94\%. Among training data and training strategy, using \change{a} fine tuned MVFCNN has the most impact of 32.48 \scndchange{percent points} improvement compared \change{to MVFCNNs with unbalanced data that are trained from scratch}. In Table \ref{tab:3}, the accuracy of the methods using LOM images have been provided which \change{are} able to achieve \change{up to} 70.14\% accuracy with \change{a} similar configuration \change{as} the best results using SEM images. \change{ These results indicate that, by considering the available dataset, the LOM is not as useful as the SEM because of the \scndchange{feature} size of the objects in this microstructural classification approach. \scndchange{However,} there is still a potential to increase the LOM accuracy by taking into account \scndchange{e.g., }color etchings like Klemm or contrasting described in the literature \cite{vander1}}. In \change{Table} \ref{tab:confusionmat}, \change{the} confusion matrix of \change{the} MVFCNN approach (fine tuned with balanced and augmented SEM training data) as the best performing method without matrix (ferrite)\change{, is} shown. In this matrix, missed objects in segmentation step \change{are} not taken into account (\#48 objects). That is why the overall accuracy is 95.23\% \change{(see Table \ref{tab:confusionmat})} which is slightly higher than the best classification accuracy of 93.94\%, shown in Table \ref{tab:1}. \change{The} confusion matrix shows the number of samples for each class predicted by the system. Recall and precision numbers show \change{the} correct classification percentage of actual classes and predictions, respectively \scndchange{ e.g.,} the network has classified 1190 martensite objects correctly which is 94.97\% of the whole martensitic objects, \change{Table \ref{tab:confusionmat}, column 1. On the other hand, the network has misclassified 24 and 39 martensitic objects as tempered martensite and bainite, respectively. This performance gives 94.97\% recall rate and 99.08\% precision rate.} \change{The dataset is heavily unbalanced, especifically for tempered matersitic samples. Even though there is one training image for tempered martensite, it is a big image \change{for} which, after cropping and applying data augmentation techniques, the number of patches are far more than one. In our experiments without data augmentation, we could not get a high precision rate for tempered martensitic samples. However, after having used data augmentation techniques, the precision rate for tempered martensite was improved dramatically.} In Table \ref{tab:4}, the results of \change{the} pixel-wise semantic segmentation with different configurations \change{are} presented. In this table, as expected, \change{the} FCNN method for pixel-wise segmentation step can achieve the best results. It can achieve \change{a} pixel accuracy of 93.92\%, \change{a} mean pixel accuracy of 76.70\%, \change{a} mean intersection over union of 67.84\% and \change{a} frequency weighted intersection over union of 88.81\%. In Table \ref{tab:5}, the same configurations have been evaluated pixel-wise for accuracy of each class. \change{The} matrix has the highest pixel accuracy of 94.22\% as expected. And bainite has the lowest pixel accuracy of 37.32\%. \change{The} matrix is present in all  examples which is the reason \change{why} the network has learned its structure well \change{ - although it has to be mentioned that so far, the grain boundaries are neglected.} However, bainite has small objects and also one of two bainite test images is the \change{transformed} bainite sample which \change{the} network has not seen \change{what leads} to \change{a} poor performance in this class. Surprisingly, LOM can achieve \change{a} pixel accuracy of matrix with 94.11\% \change{which is comparable to} the SEM images.

\begin{figure}[!htb]
\centering
        \begin{subfigure}[b]{0.7\textwidth}               
	\includegraphics[width=1\linewidth]{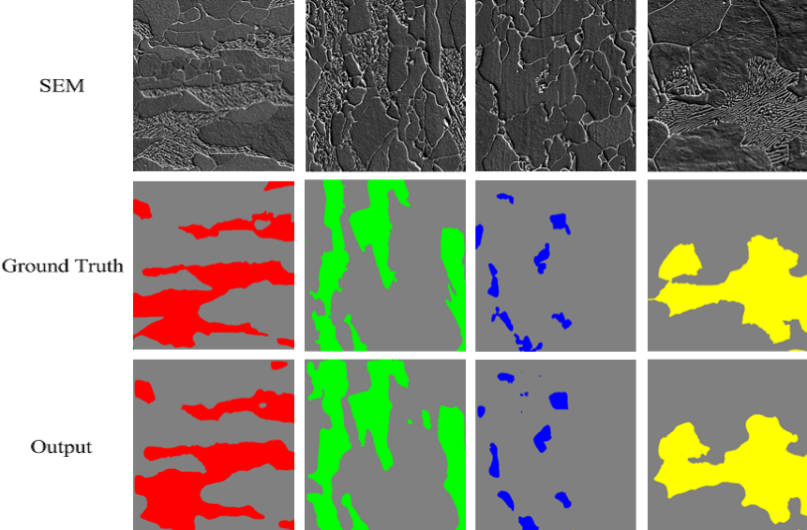}
		\caption{Success cases}
		\label{fig:success}
	\end{subfigure}
        \begin{subfigure}[b]{0.7\textwidth}
	\includegraphics[width=1\linewidth]{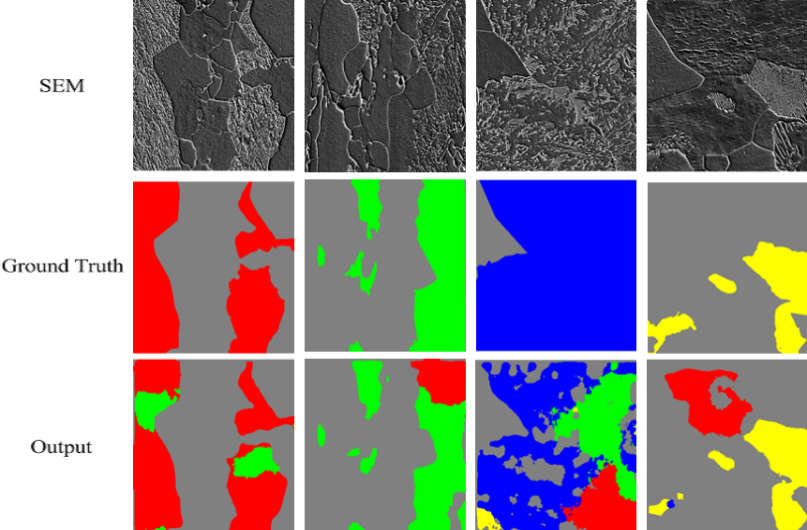}
		\caption{Failure cases}
		\label{fig:failure}
	\end{subfigure}
	\caption{a)Examples of a)successful \change{and} b)failure cases in SEM segmentation using the best MVFCNN approach configuration. The ground truth colors of martensite, tempered martensite, bainite and pearlite are red, green, blue and yellow respectively.}
	\label{fig:results}
\end{figure}

\begin{figure}[!htb]
	
	\begin{subfigure}{0.12\textwidth}
	\centering
	\includegraphics[width=0.8\textwidth]{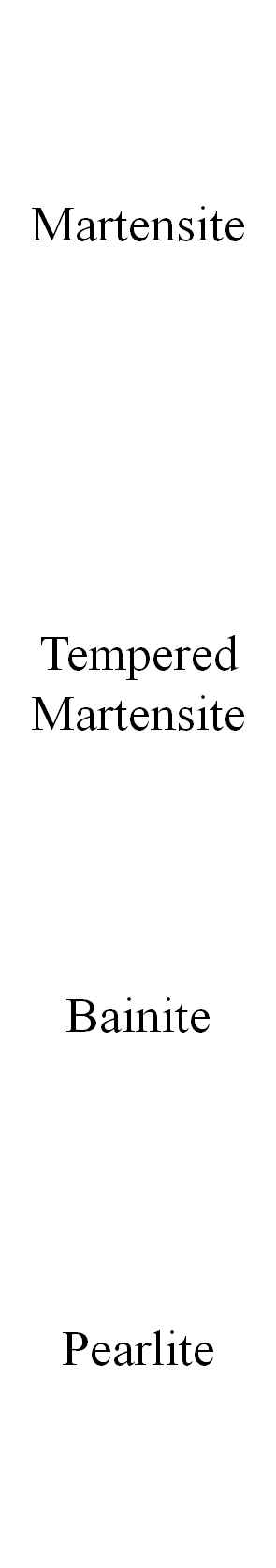}
	\end{subfigure}
	\begin{subfigure}{0.28\textwidth}
		\includegraphics[width=0.8\textwidth]{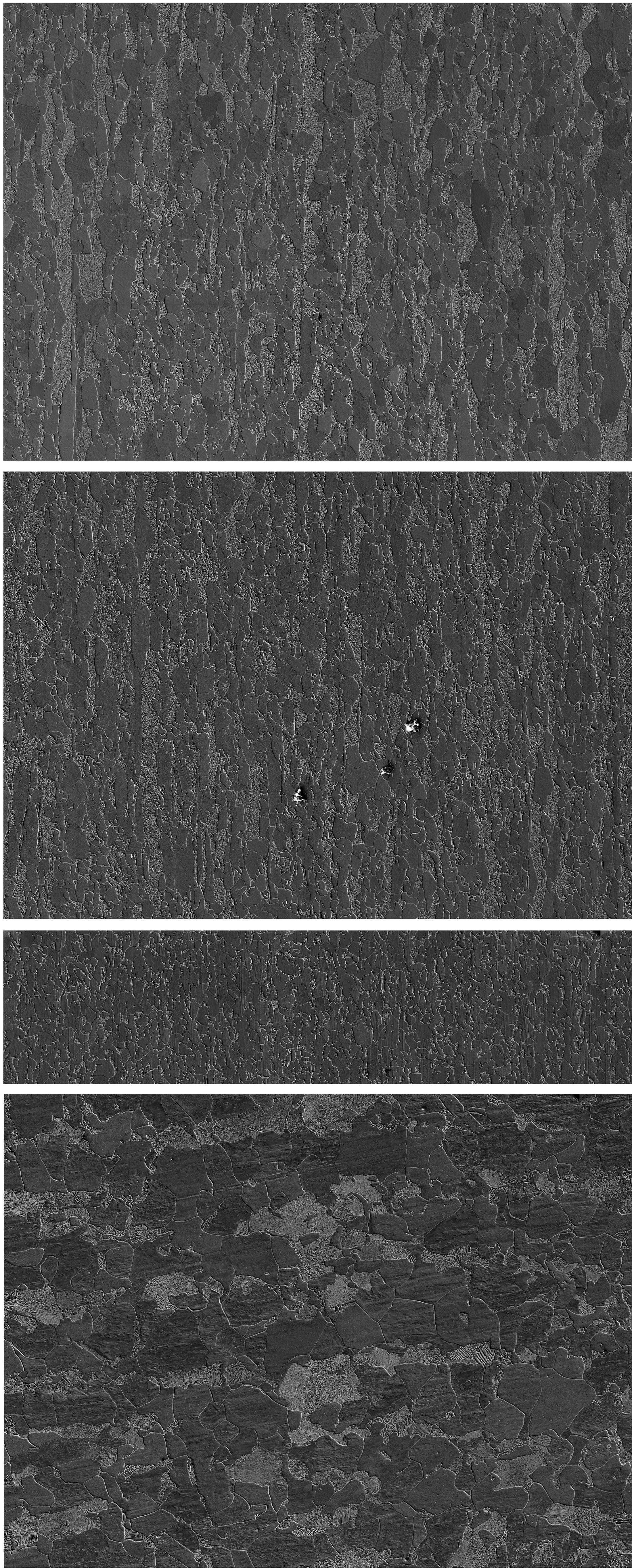}
		\caption{Original images}
	\end{subfigure}
	\begin{subfigure}{0.28\textwidth}
		\includegraphics[width=0.8\textwidth]{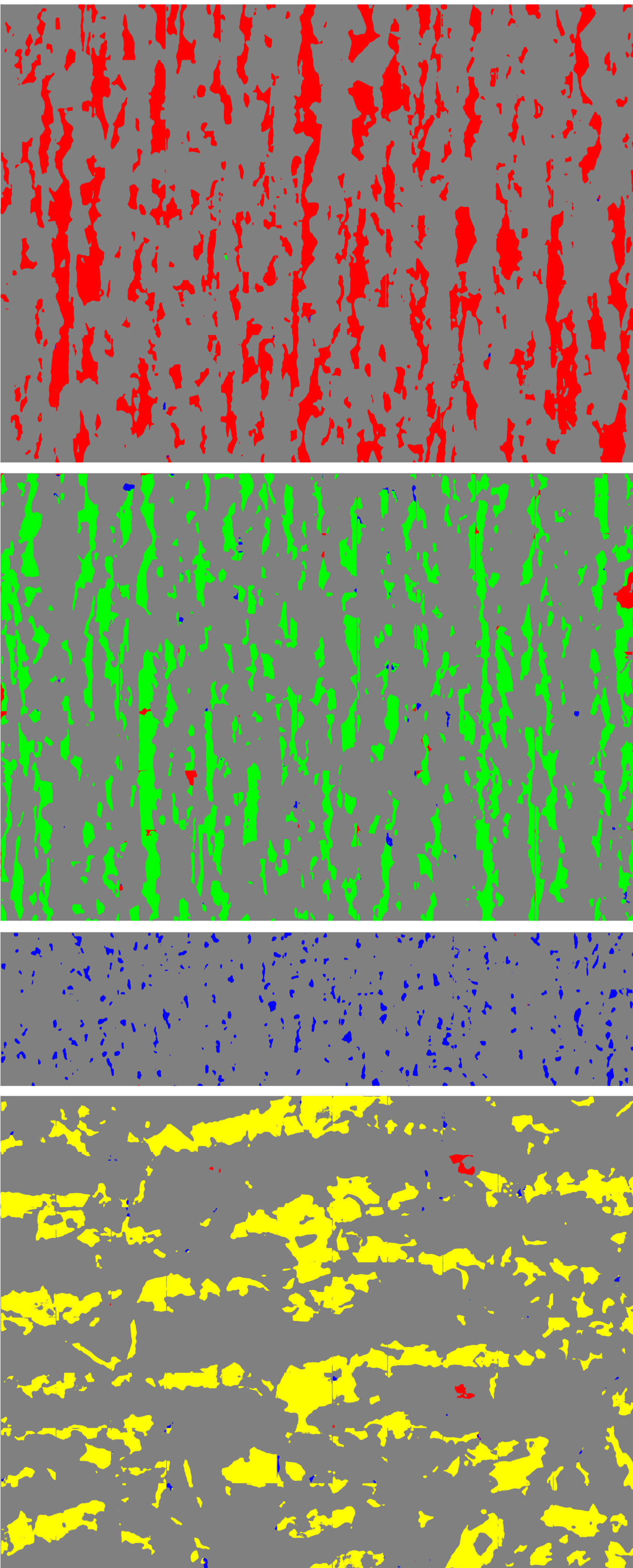}
		\caption{Stitched segmented patches}
	\end{subfigure} 
	\begin{subfigure}{0.28\textwidth}
		\includegraphics[width=0.8\textwidth]{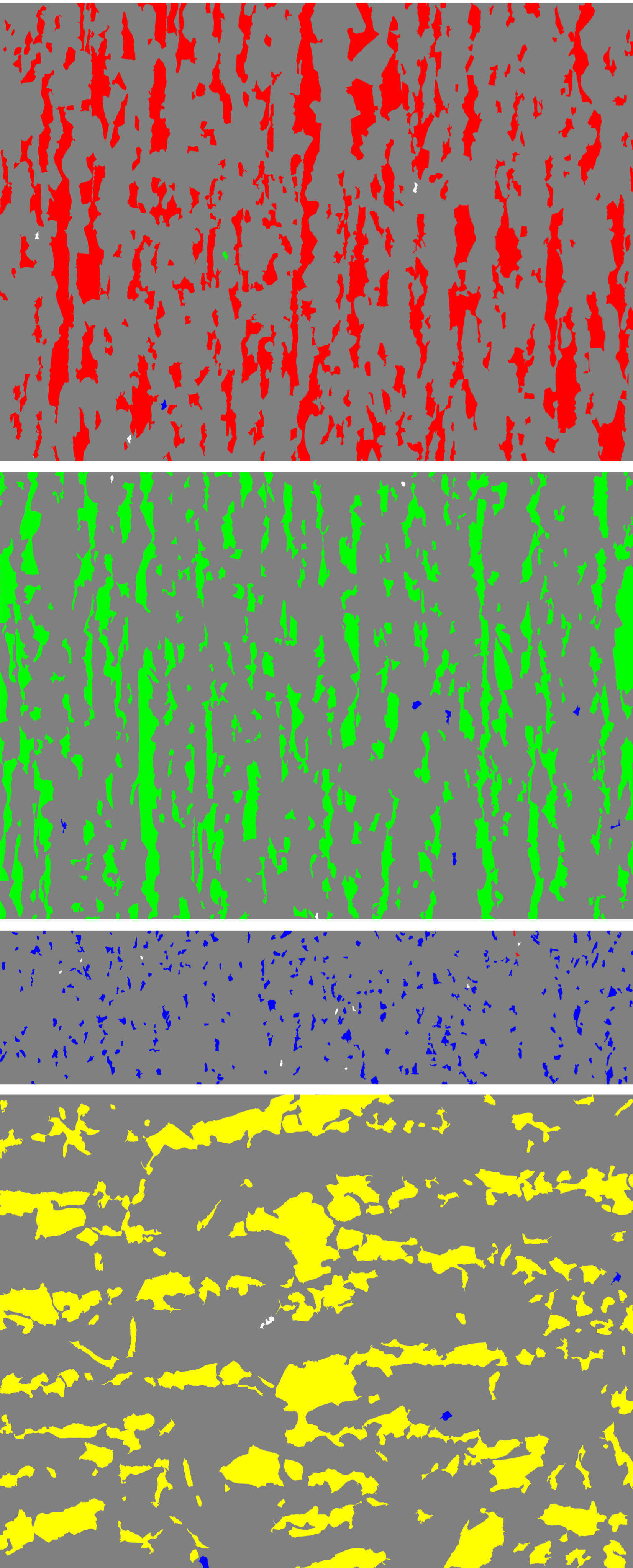}
		\caption{Classification after max-voting}
	\end{subfigure} 
	\caption{ Examples of applying \change{a} max-voting scheme to stitched \change{and segmented} SEM patches for different microstructure classes. The ground truth colors of martensite, tempered martensite, bainite and pearlite are red, green, blue, and yellow, respectively. Not-segmented objects in the segmentation phase are shown with white color.}
	\label{fig:applmax}
\end{figure}

In Figure \ref{fig:success}, some successful examples of SEM segmentation using FCNN networks, trained with balanced and augmented training data \change{are} shown \change{next to} some failure cases in Figure \ref{fig:failure}. Regarding \change{the} failure case of bainite, it should be noted that \change{the network was trained using isothermally transformed bainitic specimens} which showed no failure cases in \change{the} test set. The final results of \change{the microstructural} classification by stitching and applying \change{a} max-voting scheme over the segmented patches \change{are} depicted in Figure \ref{fig:applmax}. The final results in this figure show that most of \change{the} objects in each microstructure image \change{are} classified correctly. If we consider the classification of the \change{overall} microstructure \change{of each} image, all of \change{the ten} test images are classified correctly.

\begin{table}[!htb]
	\centering
	\caption{The effect of \change{fine tuning} and data augmentation techniques using \change{the} MVFCNN approach \change{are} depicted. The results show \change{that fine tuning} together with data augmentation achieves the best result. However, the effect of data augmentation is not significant.}
	\label{tab:2}
	\begin{tabular}{cccc}
    \toprule
Network Architecture & 
 Training Data &
Training Strategy&
Test Accuracy \\ \midrule
		MVFCNN& Unbalanced SEM&from scratch & 55.50\%\\ 
		MVFCNN& Unbalanced SEM & fine tune MVFCNN& 87.98\%\\ 
		MVFCNN& Balanced SEM& fine tune MVFCNN & 90.97\%\\ 
		MVFCNN& Balanced and augmented SEM & fine tune MVFCNN & \textbf{93.94\%}\\ \bottomrule
	\end{tabular}
\end{table}
\begin{table}[!htb]
	\centering
	\caption{The results of the same experiments with LOM images are provided in this table. The results show inferior performance using LOM \change{instead of} SEM images.}
	\label{tab:3}
	\begin{tabular}{lcccc}
    \toprule
		Method & \begin{tabular}[c]{@{}c@{}}Input\\ Data\end{tabular}&\change{Data}&\begin{tabular}[c]{@{}c@{}}Training\\ Strategy\end{tabular} & Test Accuracy \\ \midrule
        
\begin{tabular}[c]{@{}c@{}}CIFARNet\end{tabular} & cropped object  & LOM& from scratch &51,27\% \\
        
\begin{tabular}[c]{@{}c@{}}VGG-19- DeCAF-\\ RBF-kernel SVM\end{tabular} & cropped object & LOM & -- & 56.56\% \\
        
\begin{tabular}[c]{@{}c@{}}VGG-16\end{tabular} & cropped object &  LOM & \change{fine tuning} & 60,02\%  \\

MVFCNN & cropped patch &  LOM & \change{fine tuning} & 70.14\% \\
     
MVFCNN & cropped patch & SEM & \change{fine tuning} & \textbf{93.94\%} \\
\bottomrule  
	\end{tabular}
\end{table}

\begin{table}[!htb]
\begin{center}
	\caption{Confusion Matrix of the best MVFCNN approach. \change{The} matrix shows the number of samples for each class predicted by the system. Due to \change{an} unbalanced multi-class problem, percentage numbers for each class show normalized recall rates. Note: \#48 not-segmented objects have not been considered. Overall accuracy  is 93.94\%.}
	\label{tab:confusionmat}
	\begin{tabular}{c|c|c|c|c|c|c|}

   \multicolumn{2}{c}{}& \multicolumn{4}{c}{\centering  \textbf{Actual Class Labels}}&\multicolumn{1}{c}{}\\
\cline{3-7}
 \multicolumn{2}{c|}{}&
 {Martensite}
&
{\begin{tabular}[c]{@{}c@{}} Temp.\\ Martensite \end{tabular} }&
{Bainite}&
{Pearlite}& \textbf{Class Precision}\\
\cline{2-7}
        
\multirow{5}{*}{\rotatebox{90}{ \textbf{Predicted Class Labels}}}&Martensite&
\begin{tabular}[c]{@{}c@{}}1190\\94.97\%\end{tabular}&
\begin{tabular}[c]{@{}c@{}}0\\0.00\%\end{tabular}&
\begin{tabular}[c]{@{}c@{}}11\\3.19\%\end{tabular}&
\begin{tabular}[c]{@{}c@{}}0\\0.00\%\end{tabular}&99.08\%\\
\cline{2-7}

&\begin{tabular}[c]{@{}c@{}} Temp.\\ Martensite \end{tabular}&
\begin{tabular}[c]{@{}c@{}}24\\1.92\%\end{tabular}&
\begin{tabular}[c]{@{}c@{}}268\\97.81\%\end{tabular}&
\begin{tabular}[c]{@{}c@{}}0\\0.00\%\end{tabular}&
\begin{tabular}[c]{@{}c@{}}0\\0.00\%\end{tabular}&91.78\%\\\cline{2-7}

&Bainite&
\begin{tabular}[c]{@{}c@{}}39\\3.11\%\end{tabular}&
\begin{tabular}[c]{@{}c@{}}6\\2.19\%\end{tabular}&
\begin{tabular}[c]{@{}c@{}}325\\94.20\%\end{tabular}&
\begin{tabular}[c]{@{}c@{}}16\\4.80\%\end{tabular}&84.19\%\\\cline{2-7}

&Pearlite&
\begin{tabular}[c]{@{}c@{}}0\\0.00\%\end{tabular}&
\begin{tabular}[c]{@{}c@{}}0\\0.00\%\end{tabular}&
\begin{tabular}[c]{@{}c@{}}9\\2.61\%\end{tabular}&
\begin{tabular}[c]{@{}c@{}}317\\95.20\%\end{tabular}&97.23\%\\
\cline{2-7}
\cline{2-7}

& \textbf{Class Recall} & 94.97\% & 97.81\% & 94.20\% & 95.19\% & \begin{tabular}[c]{@{}c@{}}\textbf{Total Accuracy}\\95.23\%\end{tabular}\\\cline{2-7}
		\end{tabular}
    \end{center}
\end{table}

\begin{table}[!htb]
	\centering
	\caption{Evaluation of \change{the} segmentation-based approach using FCNNs with different training data and training strategies. As the results show, using SEM images, fine tuned networks and augmented data, \change{the} best performance can be achieved. However, LOM has \change{an} inferior performance compared \change{to} SEM.}
	\label{tab:4}
	\begin{tabular}{ccccccccc}
    \toprule
		Network&\change{Data}&Balanced&Augmented&
        \begin{tabular}[c]{@{}c@{}}Training\\Strategy\end{tabular}&
		\begin{tabular}[c]{@{}c@{}}pixel\\acc.\end{tabular}&
		\begin{tabular}[c]{@{}c@{}}mean\\acc.\end{tabular}&
		\begin{tabular}[c]{@{}c@{}}Mean\\IU\end{tabular}&
        \begin{tabular}[c]{@{}c@{}}f.w.\\IU\end{tabular}\\\midrule
		FCNN&SEM&-&-&scratch&80.84&33.99&24.25&71.74 \\
		FCNN&SEM&-&-&\change{fine tuning}&92.01&76.26&68.26&86.80 \\
		FCNN&SEM&\checkmark&-&\change{fine tuning}&92.11&79.63&66.27&86.91\\
		FCNN&SEM&\checkmark&\checkmark&\change{fine tuning}&\textbf{93.92}& \textbf{76.70} &\textbf{67.84}&\textbf{88.81}\\
        FCNN&LOM&\checkmark&\checkmark&\change{fine tuning}&88.27&54.01&45.43&81.66\\
        \bottomrule
	\end{tabular}
\end{table}

\begin{table}[!htb]
	\centering
	\caption{Pixel-wise accuracy evaluation of \change{the} segmentation approach using FCNNs for each microstructure class. \change{The} matrix has the highest and bainite the lowest pixel-wise accuracy.}
	\label{tab:5}
	\begin{tabularx}{\textwidth}{c@{\extracolsep{17pt}} c@{\extracolsep{17pt}}c@{\extracolsep{17pt}}c@{\extracolsep{17pt}}c@{\extracolsep{17pt}}c@{\extracolsep{17pt}}c@{\extracolsep{17pt}}c@{\extracolsep{17pt}}c@{\extracolsep{17pt}}c@{\extracolsep{17pt}}c}
    \toprule
		Network & 
		\change{Data}&
		Balanced&
        Augmented&
		{\begin{tabular}[c]{@{}c@{}}Training\\Strategy\end{tabular}}&
		Matrix&
		Marten.&
		{\begin{tabular}[c]{@{}c@{}}Temp.\\ Marten.\end{tabular}}&
		Bain.&
        Pear.\\\midrule
        
		FCNN&SEM&-&-&scratch&86.47&36.08&0&0&0\\
		FCNN&SEM&-&-&\change{fine tuning}&92.03&77.25&71.43&22.05&63.27\\
		FCNN&SEM&\checkmark&-&\change{fine tuning}&92.47&74.71&57.44&33.25&69.65\\
		FCNN&SEM&\checkmark&\checkmark&\change{fine tuning}&\textbf{94.22}& 	\textbf{79.85}&\textbf{72.62}&\textbf{37.32}&\textbf{70.46}\\
		FCNN&LOM&\checkmark&\checkmark&\change{fine tuning}&94.11&50.11&58.36&5.21& 19.35\\
        \bottomrule
	\end{tabularx}
\end{table}

\begin{figure}[!htb]
        \begin{subfigure}[b]{0.67\textwidth}
                \centering
	\includegraphics[width=0.8\linewidth]{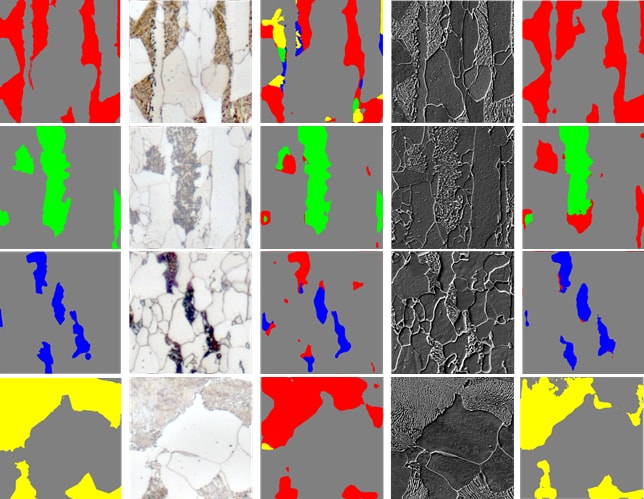}
		\caption{Comparison of SEM and LOM pixel-wise semantic segmentation}
		\label{fig:semlomcompare}
	\end{subfigure}
        \begin{subfigure}[b]{0.255\textwidth}
                \centering
	\includegraphics[width=0.8\linewidth]{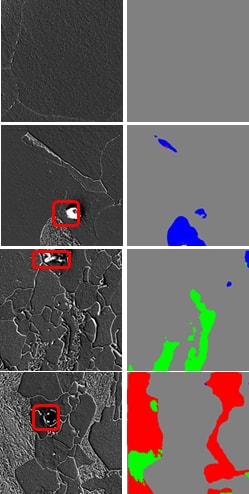}
		\caption{Model robustness to noise}
		\label{fig:noisypatch}
	\end{subfigure}
	\caption{(a) Comparison of LOM and SEM segmentation using FCNN network. Columns from left to right: Ground truth, LOM, LOM segmentation, SEM and SEM segmentation. (b) Noisy patches segmentation. First top row shows ferrite and the rests is noise. \change{The ground truth colors of martensite, tempered martensite, bainite and pearlite are red, green, blue, and yellow, respectively.}}
    \label{semlomnoise}
\end{figure}

To compare segmentation performances between segmented LOM and SEM images, in Figure \ref{fig:semlomcompare}, segmentation of patches of four microstructures using LOM and SEM images \change{by the} FCNN approach, trained with balanced and augmented training data have been shown. In most cases, SEM segmentation has a better accuracy, but it is interesting that the network can \change{still} learn correct pixel-wise classification from a low-resolution LOM images \change{compared to SEM} which suffer from different illumination and stitching artifacts. Figure \ref{fig:noisypatch} shows a few noisy sample patches in SEM images with \change{the} corresponding segmentations. Noise \change{originates} mostly from dirt\change{, dust and sample preparation}. \change{The} matrix (ferrite) segmentation is  considered as background \change{which is} also presented showing a very good robustness.

\section{Discussion}
\paragraph{Classification using object-based CNNs:} Based on Table \ref{tab:1}, all of \change{the} Deep Learning methods outperform the state-of-the-art method which proves our motivation regarding using learned high-level features rather than engineered features. The results also indicate \change{that} deeper networks show better results than shallower ones \change{ i.e.} depth matters. Even with features extracted from pre-trained VGG19, which are classified with SVM, one can achieve comparable performances with the results of trained CIFARNET and VGG16. However, training VGG16 on \change{the} dataset \change{used in this work} makes features more informative and discriminative as network parameters can learn the pattern in this dataset. The surprising performance of \change{the} MVFCNN method indicates that \change{the} performance of object-based CNNs is negatively influenced by the image resizing step. In other words, we observe that resizing \change{distorts} the texture inside objects, hampering the accurate differentiation of objects bigger or smaller than \change{the} input size of \change{the} network. \change{In another approach, we }\scndchange{split} big objects into 224x224 px objects \scndchange{which led to a} higher performance which supports our \change{assumption}. However, this approach made the system less practical and introduces a hyper parameter into the system to choose the best split size.

\paragraph{Classification using MVFCNNs:} The results in Table \ref{tab:1} shows \change{that a} MVFCNN approach using SEM images achieves \change{a very high} performance. It indicates that \change{a} classification using pixel-wise segmentation is more efficient and accurate compared \change{to} object-based CNN methods and significantly better than hand-crafted features. Resizing is no longer required in this method and therefore does not have an impact on \change{the performance of the }MVFCNN. 

The confusion matrix in \change{Table} \ref{tab:confusionmat} shows that the network still produces some misclassification of martensite objects due to the confusion of martensite with tempered martensite and bainite which have similar textures -- unlike \change{the texture of} pearlite's texture which is easy to distinguish. All \change{wrongly} classified bainite objects belong to \change{the ``transformed''} bainite sample in the test set which is not present in the \change{training} set. It is impressive that in this condition the network is still able to classify more than half of \change{the} objects in \change{the ``transformed''} bainite sample correctly. The achieved high accuracy indicates that considering each pixel and taking into account its neighboring pixels plays a crucial role in the correct classification of objects. 

Based on the results of Table \ref{tab:2}, data augmentation improves the performance by 2 \scndchange{percent points} which is not considerable. One possible reason for this phenomenon is that different rotations of textures inside objects are already present in the dataset before data augmentation which the network has already seen and learned. For example, in \change{the} pearlite microstructure, there are many cases that by rotating a patch, the resulting augmented patch still contains the previous orientations before the rotation. \change{The results presented in }Table \ref{tab:3} confirm our expectation that LOM images will perform poorly compared \change{to} SEM due to \change{lower resolution}. The artifacts due to stitching as well as the different illumination \change{and preparation residues and scratches} in LOM images degrade classification accuracy. The results in Table \ref{tab:3} have the same trend compared with Table \ref{tab:1} and confirm our findings. Table \ref{tab:4} shows \change{that} using pixel-wise classification using MVFCNNs, one can achieve \change{a} high performance of 93.94\% accuracy \change{de}spite of more complex microstructural textures. These results also show \change{that} the better \change{the} pixel-wise segmentation criteria are, the better \change{the microstructural} classification will be.

Regarding Table \ref{tab:5}, the results show how important \change{fine tuning} is when working with little training data. Without \change{fine tuning}, tempered martensite, bainite and pearlite microstructures could have not been segmented at all. Low performance in bainite class even by doing data augmentation could be due to the fact that bainite objects compared \change{to} other classes are a lot smaller. In the dataset, there are quite a few tiny areas belonging to matrix class which are not completely plain and contain some textures similar to those in bainite. As as result, balancing and augmenting the dataset makes the network decide that those small objects are more likely matrix than bainite. This assumption is verified by observing the bainite test image that pixels are either classified as bainite or as matrix. Another reason is also presence of \change{isothermally transformed} bainite which has different appearance with bainite \change{of the other sample} and network was not trained on that. More training data, specially for bainite class \change{according to \cite{zajac}} would help to decrease the miss-classified bainite objects in the segmentation step. \change{This is in process for the next generation although it is very hard to find enough samples with a sufficient number of objects - especially for upper and lower bainite - which can be separated from a matrix to define the ground truth and to have a sufficiently good training dataset. Moreover, first  efforts are heading in the  direction of producing lab melts which are then analyzed by dilatometry to further enhance the division of the classes. Also the ferrite including its grain boundaries will be added in a next iteration as additional class.} Regarding noise robustness, Figure \ref{fig:noisypatch} shows strong robustness of the system with different types of noises. We also noted that the system is rotation \change{invariant} which means similar patches with different rotation angles have the same system response. It is worth mentioning that MVFCNNs can also have input images smaller than 1000x1000 px. However, in input image smaller than 100x100 px the response is for the same area worse than \change{has} 1000x1000 px. 

\section{Conclusion}
This work demonstrates the feasibility of an effective steel \change{microstructural} classification using Deep Learning methods without a need of separate segmentation and feature extraction. \change{We} performed a pixel-wise microstructural segmentation using a trained FCNN network followed by a max-voting scheme. \change{The observed strong improvements in classification performance over the prior state of the art} confirm our idea of leveraging the raw data input for training Deep Learning-based classification systems. Besides the high accuracy result, we are able to achieve a very fast prediction.

We found that resizing objects directly as input to object-based CNNs can eliminate discriminative texture information relevant to different \change{microstructural} classes. In contrast, MVFCNN approach does not have this problem and it is independent of the size of objects. Data augmentation was considered for further performance improvements. Furthermore, we found that rotating the SEM images does not introduce considerable new information and therefore the performance is not significantly improved. We conclude that pixel-wise segmentation using Fully Convolutional Neural Networks is an effective and robust way of determining the distribution and size of different microstructures when these networks are trained end-to-end.

\section{Acknowledgements}
    We are grateful to NVIDIA company for granting us with a GPU.
    

\section{Additional information}
The authors declare no competing financial interests.

\bibliography{smkdp}
\end{document}